\documentclass[letterpaper]{article} 
\usepackage{aaai25}  
\usepackage{times}  
\usepackage{helvet}  
\usepackage{courier}  
\usepackage[hyphens]{url}  
\usepackage{subcaption}
\usepackage{float}
\usepackage{mathrsfs}
\usepackage{tabularx}
\usepackage{xcolor}

\usepackage{graphicx} 
\usepackage{natbib}  
\usepackage{caption} 
\usepackage{booktabs}
\usepackage{algorithm}
\usepackage{algorithmic}
\usepackage{newfloat}
\usepackage{listings}
\usepackage{pifont}
\usepackage{amsmath,amssymb}
\usepackage{xcolor}
\usepackage{pifont}

\newcommand{\tick}{\textcolor{red}{\ding{51}}} 
\newcommand{\cross}{\ding{55}} 

\DeclareCaptionStyle{ruled}{labelfont=normalfont,labelsep=colon,strut=off} 
\floatstyle{ruled}
\newfloat{listing}{tb}{lst}{}
\floatname{listing}{Listing}
\title{Spatial Distribution-Shift Aware Knowledge-Guided Machine Learning\vspace{-1.25em}}
\author {
    Arun Sharma\textsuperscript{\rm 1},
    Majid Farhadloo\textsuperscript{\rm 1},
    Mingzhou Yang\textsuperscript{\rm 1},
    Ruolei Zeng\textsuperscript{\rm 1},
    Subhankar Ghosh\textsuperscript{\rm 1},
    Shashi Shekhar\textsuperscript{\rm 1}
}
\affiliations {
    \textsuperscript{\rm 1}Department of Computer Science, University of Minnesota, Twin Cities, USA\\

    \{sharm485, farha043, yang7492, zeng0208, ghosh117, shekhar\}@umn.edu
}
\usepackage{bibentry}
\begin{document}

\maketitle
\vspace{-5em}
\begin{abstract}
Given inputs of diverse soil characteristics and climate data gathered from various regions, we aimed to build a model to predict accurate land emissions. The problem is important since accurate quantification of the carbon cycle in agroecosystems is crucial for mitigating climate change and ensuring sustainable food production. Predicting accurate land emissions is challenging since calibrating the heterogeneous nature of soil properties, moisture, and environmental conditions is hard at decision-relevant scales. Traditional approaches do not adequately estimate land emissions due to location-independent parameters failing to leverage the spatial heterogeneity and also require large datasets. To overcome these limitations, we proposed Spatial Distribution-Shift Aware Knowledge-Guided Machine Learning (SDSA-KGML), which leverages location-dependent parameters that account for significant spatial heterogeneity in soil moisture from multiple sites within the same region. Experimental results demonstrate that SDSA-KGML models achieve higher local accuracy for the specified states in the Midwest Region.
\end{abstract}
\vspace{-1.5em}
\section{Introduction}
\begin{figure}[h]
\centering
\begin{subfigure}[b]{0.40\linewidth} 
    \centering
    \includegraphics[width=\linewidth]{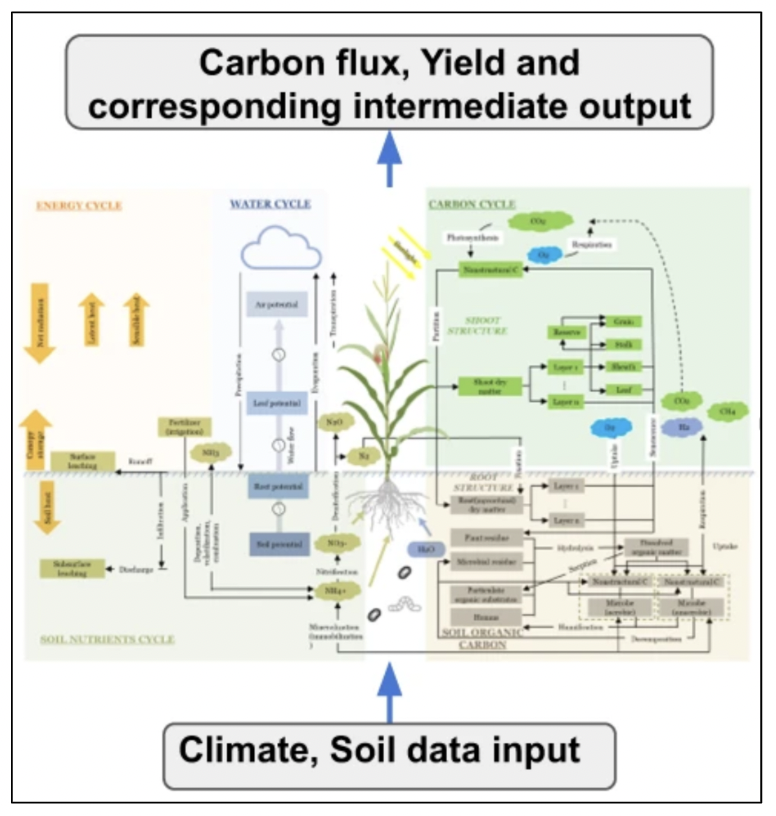}
    \caption{Input}
    \label{fig:subfig1}
\end{subfigure}
\hfill 
\begin{subfigure}[b]{0.59\linewidth} 
    \centering
    \includegraphics[width=\linewidth]{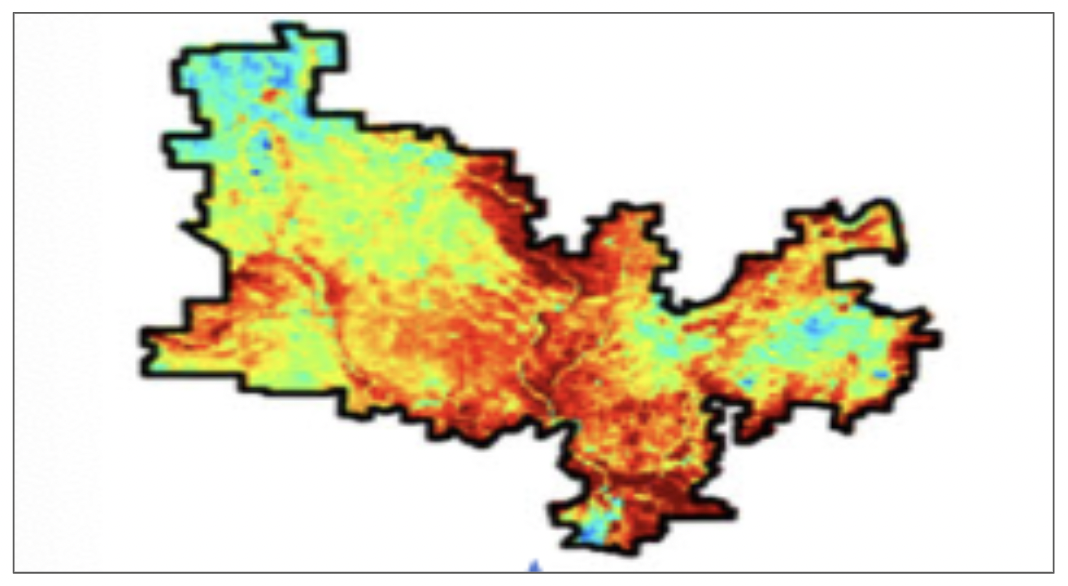}
    \caption{Output}
    \label{fig:subfig2}
\end{subfigure}
\caption{Problem Statement \cite{liu2024knowledge}}
\label{fig1}
\vspace{-2em} 
\end{figure}

Given inputs of diverse types of information, including climate data, soil characteristics, and Gross Primary Productivity (GPP) gathered from various regions, we aimed to build a model to predict land emissions. These emissions include the amount of carbon dioxide released by vegetation through Autotrophic Respiration (Ra) and the carbon dioxide released by soil microorganisms \cite{janssens2007spatial} during the decomposition of organic matter, termed Heterotrophic Respiration (Rh), for various regions \cite{liu2024knowledge}. Figure \ref{fig1} shows estimated land emissions generated by a knowledge-guided machine learning model using climate data and soil characteristics \cite{liu2024knowledge}.

Land emissions estimations, such as carbon dioxide, methane, and nitrogen oxides, are important for sustainable agriculture. They also support climate change mitigation, optimized crop management, and informed decision-making to sustain food production. Given these initiatives, it is essential to establish reliable and scalable methods for accurately quantifying carbon sequestration at the field level. This will help evaluate its impact on climate mitigation and ensure that farmers' mitigation efforts are compensated fairly and precisely. The problem is challenging since calibrating the heterogeneous nature of soil properties, moisture, and environmental conditions results in lower accuracy. In addition, such models requires large dataset \cite{2022a,2022b}.

\textbf{Limitations of Related Work:} Knowledge Guided Machine Learning (KGML) \cite{karpatne2017theory,karpatne2022knowledge} has shown promise in modeling earth systems governed by established equations \cite{trivedi2020knowledge}. For example, KGML-ag \cite{liu2022kgml} combines process-based models to improve agroecosystem dynamics predictions, including emission estimates, while \citet{liu2024knowledge} refines carbon cycle estimations by integrating process-based models. However grounded in principles like mass and energy conservation are commonly used for predicting land emissions but struggle in regions with high spatial heterogeneity \cite{gupta2021spatial} and require large datasets \cite{2022c}. In this work, we propose a region-based knowledge-guided machine learning framework to extract more precise and meaningful land emission estimates for precision agriculture and agroecosystems.

\textbf{Contributions:} This paper makes three contributions:
\begin{itemize}
    \item We introduced a taxonomy based for spatial variability to study the impact of location-based model parameters.
    \item We briefly introduce the Spatial Distribution-Shift Aware Knowledge-Guided Machine Learning (SDSA-KGML) which consider location-dependent parameters.
    \item We validated SDSA-KGML achieving higher local accuracy across Illinois, Iowa, and Indianna.
\end{itemize}




\begin{table*}[t]
\small
\centering
\caption{Spatial Variability Awareness Levels based on Location Dependent (\tick) and Location Independent (\cross)}
\label{tab:table1}
\begin{tabular}{ccccccc}
\toprule
Level & Taxonomy & Example & Inputs ($x$) & Outputs ($y$) & Parameters ($\theta$)\\
\midrule
1 & One Size Fit All & Data-Driven Models & \cross & \cross & \cross\\
2 & Spatial Explicit & KGML-Ag \cite{liu2024knowledge}) & \tick & \tick & \cross \\
3 & Spatial Variability-Aware & \textcolor{red}{\textbf{Proposed Approach}} & \tick & \tick & \tick\\
\bottomrule
\end{tabular}
\end{table*}

\vspace{-1em}
\section{Proposed Approach}\label{sec2}
\textbf{Taxonomy:} Land emission forecasting has been widely studied using methods such as decision trees \cite{adegun2023state}, random forests \cite{fang2018modeling,ardeshir2014gis}, and neural networks \cite{feng2023survey}. While these data-driven models handle geographic heterogeneity, they often suffer from overfitting, limiting their ability to generalize to unseen data. Table \ref{tab:table1} we present a taxonomy of ML models based on their level of spatial variability awareness. Data-driven models are one-size-fits-all as they do not consider location whereas methods like KGML-ag\cite{liu2024knowledge}) are spatial explicit but are location-independent. By contrast, the proposed approach (SDSA-KGML) incorporates location dependence into the model itself.


\begin{figure}[ht!]
    \centering
    \includegraphics[width=0.85\linewidth]{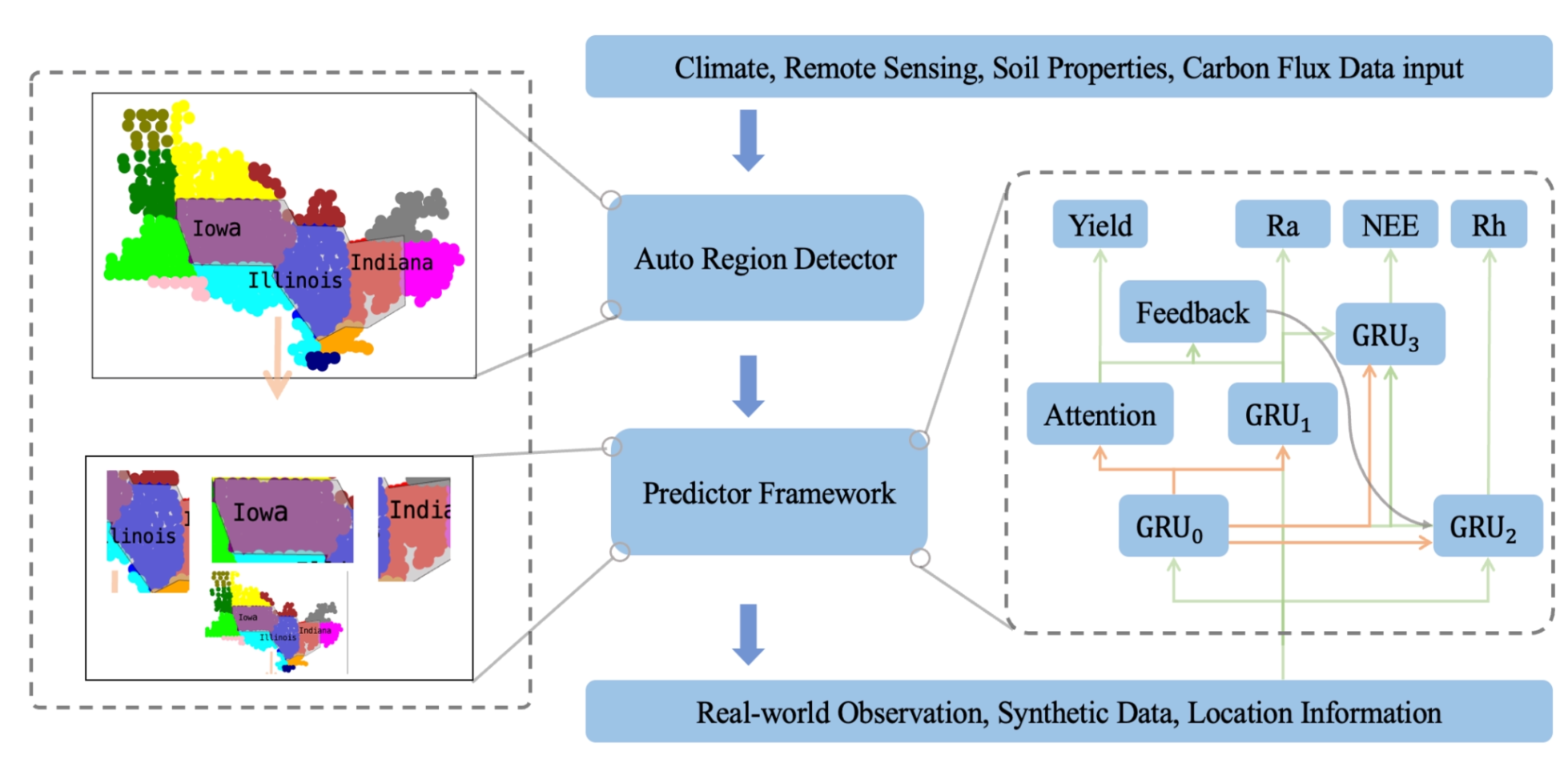} 
    \caption{Illustration of the SDSA-KGML framework.}
    \label{fig:Fig1b}
\end{figure}

\textbf{SDSA-KGML Framework:} The proposed approach leverages KGML-Ag architecture \cite{liu2024knowledge} which is pre-trained using synthetic data generated by the process-based model, enabling it to capture fundamental patterns and relationships. Figure \ref{fig:Fig1b} illustrate SDSA-KGML framework which first processes climate, remote sensing \cite{ma2019remote}, soil, and carbon flux data through an \textbf{Auto Region Detector} to extract geographic sub-area of of a Midwest Region i.e., \textit{Illinois}, \textit{Iowa}, and \textit{Indiana} which is later feeds to \textbf{Predictor Framework} using GRU layers and attention mechanisms to predict land emissions. The model is then fine-tuned with observed low-resolution crop yield data and carbon flux measurements from sparsely distributed eddy-covariance sites aided by knowledge-guided loss functions to  ensure that the target variables respond appropriately to input variables.  we employed a streamlined 5-step training protocol, selected the GRU architecture for the neural network, and incorporated regularization to mitigate overfitting.

\vspace{-.8em}
\section{Preliminary Results}\label{sec2}

\textbf{Dataset:} We integrated diverse datasets into a knowledge-guided learning framework and compared its performance against a one-size-fits all approach. The task was to model agroecosystem dynamics and carbon cycles for the Midwest region of the US. We utilized daily climate data from NLDAS-2\footnote{\url{https://ldas.gsfc.nasa.gov/nldas/nldas-2-forcing-data}}, soil characteristics from gSSURGO\footnote{\url{https://www.nrcs.usda.gov/resources/data-and-reports/gridded-soil-survey-geographic-gssurgo-database}}, annual crop yield data from National Agricultural Statistics Service (NASS), high-resolution GPP data from the Soil-Adjusted Near-Infrared Reflectance SANIRv model, and carbon flux measurements from EC flux towers in the U.S. Midwest. These datasets were preprocessed using Z-normalization, interpolation, and aggregation. The SDSA-KGML framework combines these datasets to predict carbon fluxes, crop yields, and soil organic carbon changes with precision, training based on 100 samples from Illinois, Indiana, and Iowa.

\begin{figure}[ht!]
    \centering
    \includegraphics[width=0.7\linewidth]{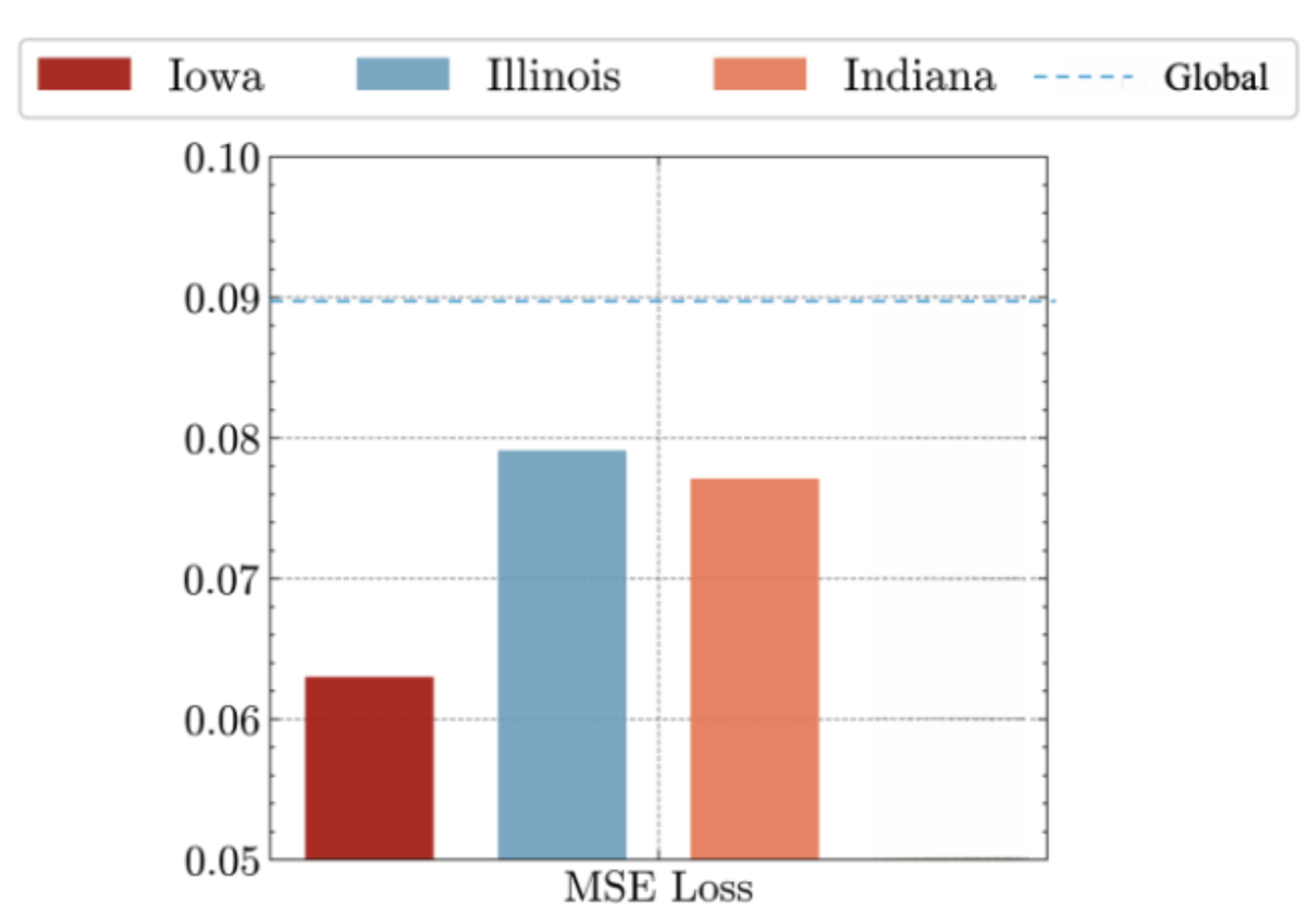} 
    \caption{MSE Loss across  Illinois, Iowa, and Indiana}
    \label{fig:fig1a}
\end{figure}

Our findings show that SDSA-KGML models trained on state-specific data outperform the global model (KGML-Ag), achieving higher R1 and R2 values and lower MSE losses. Figure \ref{fig:fig1a} highlights that models trained on Iowa, Indiana, and Illinois data have lower MSE losses when tested within these states compared to training on combined data. This demonstrates that regional data segmentation enhances geographic information integration, capturing finer nuances and improving prediction accuracy.

\vspace{-1em}
\section{Conclusion and Future Work}\label{sec5}
This study demonstrates SDSA-KGML model employs location-based parameter values to effectively capture spatial heterogeneity. Results demonstrated that models trained on data specific to individual states better accuracy than those using location-independent parameters. 

\textbf{Future Work:} We plan to add novel partitioning techniques and new datasets \cite{2015d} for detailed experiments to better capture spatial variability \cite{ghosh2024towardssigspatial, ghosh2024towardsarxiv, ghosh2024reducingarxivcoloc, ghosh2024towardscosit, ghosh2023reducing, ghosh2022towards, ghosh2017video, yang2025climate}. In addition, we also plan to incorporate Task-Adaptive Meta Learning \cite{liu2023task} for model generalization corresponding to any new tasks.

\section*{Acknowledgments}{This material is based on work supported by the USDA under Grant No. 2023-67021-39829, the National Science Foundation under Grant No. 1901099, the USDOE Office of Energy Efficiency and Renewable Energy under FOA No. DE-FOA0002044, and USDA under Grant No. 2021-51181-35861. We also thank Kim Kofolt and the Spatial Computing Research Group for their valuable comments and contributions.}
\bibliography{aaai25}

\begin{thebibliography}{24}
\providecommand{\natexlab}[1]{#1}

\bibitem[{Adegun et~al.(2023)Adegun, Fonou~Dombeu, Viriri, and Odindi}]{adegun2023state}
Adegun, A.~A.; Fonou~Dombeu, J.~V.; Viriri, S.; and Odindi, J. 2023.
\newblock State-of-the-Art deep learning methods for objects detection in remote sensing satellite images.
\newblock \emph{Sensors}, 23(13): 5849.

\bibitem[{Ardeshir et~al.(2014)Ardeshir, Zamir, Torroella, and Shah}]{ardeshir2014gis}
Ardeshir, S.; Zamir, A.~R.; Torroella, A.; and Shah, M. 2014.
\newblock GIS-assisted object detection and geospatial localization.
\newblock In \emph{Computer Vision--ECCV 2014: 13th European Conference, Zurich, Switzerland, September 6-12, 2014, Proceedings, Part VI 13}, 602--617. Springer.

\bibitem[{Fang et~al.(2018)Fang, Yu, Wang, Li, Sun, Liu, Tang, and Li}]{fang2018modeling}
Fang, S.; Yu, G.; Wang, X.; Li, S.; Sun, P.; Liu, W.; Tang, Z.; and Li, Y. 2018.
\newblock Modeling the spatial and temporal variations of soil CO2 efflux across a heterogeneous landscape.
\newblock \emph{Journal of Geophysical Research: Biogeosciences}, 123(7): 1958--1976.

\bibitem[{Feng et~al.(2023)Feng, Xu, Luo, and Yang}]{feng2023survey}
Feng, Y.; Xu, Y.; Luo, Z.; and Yang, X. 2023.
\newblock A Survey on Knowledge Distillation.
\newblock \emph{IEEE Transactions on Neural Networks and Learning Systems}.

\bibitem[{Ghosh, Sharma et~al.(2024{\natexlab{a}})}]{ghosh2024towardsarxiv}
Ghosh, S.; Sharma, A.; et~al. 2024{\natexlab{a}}.
\newblock Towards Kriging-informed Conditional Diffusion for Regional Sea-Level Data Downscaling.
\newblock \emph{arXiv preprint arXiv:2410.15628}.

\bibitem[{Ghosh, Sharma et~al.(2024{\natexlab{b}})}]{ghosh2024towardssigspatial}
Ghosh, S.; Sharma, A.; et~al. 2024{\natexlab{b}}.
\newblock Towards Kriging-informed Conditional Diffusion for Regional Sea-Level Data Downscaling: A Summary of Results.
\newblock In \emph{Proceedings of the 32nd ACM International Conference on Advances in Geographic Information Systems}, 372--383.

\bibitem[{Ghosh et~al.(2022)}]{ghosh2022towards}
Ghosh, S.; et~al. 2022.
\newblock Towards geographically robust statistically significant regional colocation pattern detection.
\newblock In \emph{Proceedings of the 5th ACM SIGSPATIAL International Workshop on GeoSpatial Simulation}, 11--20.

\bibitem[{Ghosh et~al.(2023)}]{ghosh2023reducing}
Ghosh, S.; et~al. 2023.
\newblock Reducing Uncertainty in Sea-level Rise Prediction: A Spatial-variability-aware Approach.
\newblock \emph{arXiv preprint arXiv:2310.15179}.

\bibitem[{Ghosh et~al.(2024{\natexlab{a}})}]{ghosh2024reducingarxivcoloc}
Ghosh, S.; et~al. 2024{\natexlab{a}}.
\newblock Reducing false discoveries in statistically-significant regional-colocation mining: A summary of results.
\newblock \emph{arXiv preprint arXiv:2407.02536}.

\bibitem[{Ghosh et~al.(2024{\natexlab{b}})}]{ghosh2024towardscosit}
Ghosh, S.; et~al. 2024{\natexlab{b}}.
\newblock Towards Statistically Significant Taxonomy Aware Co-Location Pattern Detection (Short Paper).
\newblock In \emph{16th International Conference on Spatial Information Theory (COSIT 2024)}, 25--1. Schloss Dagstuhl--Leibniz-Zentrum f{\"u}r Informatik.

\bibitem[{Gupta et~al.(2021)Gupta, Molnar, Xie, Knight, and Shekhar}]{gupta2021spatial}
Gupta, J.; Molnar, C.; Xie, Y.; Knight, J.; and Shekhar, S. 2021.
\newblock Spatial variability aware deep neural networks (svann): A general approach.
\newblock \emph{ACM Transactions on Intelligent Systems and Technology (TIST)}, 12(6): 1--21.

\bibitem[{Janssens et~al.(2007)Janssens, Pilegaard, Lloyd, Goulden, and Wofsy}]{janssens2007spatial}
Janssens, I.~A.; Pilegaard, K.; Lloyd, J.; Goulden, M.~L.; and Wofsy, S.~C. 2007.
\newblock Spatial and temporal variability in soil CO2 efflux in pristine South African grasslands.
\newblock \emph{Global Change Biology}, 13(6): 2060--2073.

\bibitem[{Karpatne et~al.(2017)Karpatne, Atluri, Faghmous, Steinbach, Banerjee, Ganguly, Shekhar, Samatova, and Kumar}]{karpatne2017theory}
Karpatne, A.; Atluri, G.; Faghmous, J.~H.; Steinbach, M.; Banerjee, A.; Ganguly, A.; Shekhar, S.; Samatova, N.; and Kumar, V. 2017.
\newblock Theory-guided data science: A new paradigm for scientific discovery from data.
\newblock \emph{IEEE Transactions on knowledge and data engineering}, 29(10): 2318--2331.

\bibitem[{Karpatne, Kannan, and Kumar(2022)}]{karpatne2022knowledge}
Karpatne, A.; Kannan, R.; and Kumar, V. 2022.
\newblock \emph{Knowledge guided machine learning: Accelerating discovery using scientific knowledge and data}.
\newblock CRC Press.

\bibitem[{Kumar et~al.(2015)}]{2015d}
Kumar, K.~V.; et~al. 2015.
\newblock Graph based technique for hindi text summarization.
\newblock In \emph{Information Systems Design and Intelligent Applications: Proceedings of Second International Conference INDIA 2015, Volume 1}, 301--310. Springer.

\bibitem[{Liu et~al.(2024)Liu, Zhou, Guan, Peng, Xu, Tang, Zhu, Till, Jia, Jiang et~al.}]{liu2024knowledge}
Liu, L.; Zhou, W.; Guan, K.; Peng, B.; Xu, S.; Tang, J.; Zhu, Q.; Till, J.; Jia, X.; Jiang, C.; et~al. 2024.
\newblock Knowledge-guided machine learning can improve carbon cycle quantification in agroecosystems.
\newblock \emph{Nature communications}, 15(1): 357.

\bibitem[{Liu et~al.(2022)}]{liu2022kgml}
Liu, L.; et~al. 2022.
\newblock KGML-ag: a modeling framework of knowledge-guided machine learning to simulate agroecosystems: a case study of estimating N 2 O emission using data from mesocosm experiments.
\newblock \emph{Geoscientific Model Development}, 15(7): 2839--2858.

\bibitem[{Liu et~al.(2023)Liu, Liu, Xie, Jin, and Jia}]{liu2023task}
Liu, Z.; Liu, L.; Xie, Y.; Jin, Z.; and Jia, X. 2023.
\newblock Task-adaptive meta-learning framework for advancing spatial generalizability.
\newblock In \emph{Proceedings of the AAAI Conference on Artificial Intelligence}, volume~37, 14365--14373.

\bibitem[{Ma et~al.(2019)Ma, Wu, Wang, Zhang, and Liu}]{ma2019remote}
Ma, J.; Wu, Y.; Wang, L.; Zhang, L.; and Liu, Y. 2019.
\newblock Remote sensing image scene classification with bag-of-visual-words and convolutional neural network.
\newblock \emph{Remote Sensing}, 11(9): 1107.

\bibitem[{Sharma and Shekhar(2022)}]{2022b}
Sharma, A.; and Shekhar, S. 2022.
\newblock Analyzing trajectory gaps to find possible rendezvous region.
\newblock \emph{ACM Transactions on Intelligent Systems and Technology (TIST)}, 13(3): 1--23.

\bibitem[{Sharma et~al.(2022)}]{2022a}
Sharma, A.; et~al. 2022.
\newblock Towards a tighter bound on possible-rendezvous areas: preliminary results.
\newblock In \emph{Proceedings of the 30th International Conference on Advances in Geographic Information Systems}, 1--11.

\bibitem[{Sharma et~al.(2024)}]{2022c}
Sharma, A.; et~al. 2024.
\newblock Physics-based abnormal trajectory gap detection.
\newblock \emph{ACM Transactions on Intelligent Systems and Technology}, 15(5): 1--31.

\bibitem[{Trivedi et~al.(2020)}]{trivedi2020knowledge}
Trivedi, R.; et~al. 2020.
\newblock Knowledge-guided machine learning for materials science: a review of approaches, applications, and open questions.
\newblock \emph{Journal of Materials Science}, 55(2): 718--744.

\bibitem[{Yang et~al.(2025)Yang, Jayaprakash, Ghosh, Jung, Eagon, Northrop, and Shekhar}]{yang2025climate}
Yang, M.; Jayaprakash, B.; Ghosh, S.; Jung, H.~T.; Eagon, M.; Northrop, W.~F.; and Shekhar, S. 2025.
\newblock Climate smart computing: A perspective.
\newblock \emph{Pervasive and Mobile Computing}, 102019.

\end{thebibliography}

\end{document}